\documentclass[11pt,a4paper]{article}
\usepackage[hyperref]{naaclhlt2019}
\usepackage{times}
\usepackage{latexsym}
\aclfinalcopy

\usepackage{url}

\usepackage[]{algorithm2e}
\usepackage[utf8]{inputenc} 
\usepackage[T1]{fontenc}    
\usepackage{hyperref}       
\usepackage{booktabs}       
\usepackage{amsfonts}       
\usepackage{nicefrac}       
\usepackage{microtype}      
\usepackage{multirow}
\usepackage{caption}
\usepackage{amsmath,amssymb}
\usepackage{dsfont}			
\usepackage{cases}
\usepackage{color}
\usepackage{tikz}
\usetikzlibrary{shapes,arrows,calc}
\usepackage{caption}
\usepackage{subcaption}
\usepackage{enumitem}
\usepackage{import}
\usepackage{CJKutf8}
\usepackage[super]{nth}

\newtheorem{observation}{Observation}

\title{Parallelizable Stack Long Short-Term Memory}

\author{Shuoyang Ding\quad Philipp Koehn \\
		Center for Language and Speech Processing \\
        Johns Hopkins University \\
	    {\tt \{dings, phi\}@jhu.edu}}

\date{}

\begin{document}
\maketitle
\begin{abstract}
Stack Long Short-Term Memory (StackLSTM) is useful for various applications such as parsing and string-to-tree neural machine translation, but it is also known to be notoriously difficult to parallelize for GPU training due to the fact that the computations are dependent on discrete operations. In this paper, we tackle this problem by utilizing state access patterns of StackLSTM to homogenize computations with regard to different discrete operations. Our parsing experiments show that the method scales up almost linearly with increasing batch size, and our parallelized PyTorch implementation trains significantly faster compared to the Dynet C++ implementation.
\end{abstract}

\section{Introduction}

Tree-structured representation of language has been successfully applied to various applications including dependency parsing \cite{DBLP:conf/acl/DyerBLMS15}, sentiment analysis \cite{DBLP:conf/icml/SocherLNM11} and neural machine translation \cite{DBLP:conf/acl/EriguchiTC17}. However, most of the neural network architectures used to build tree-structured representations are not able to exploit full parallelism of GPUs by minibatched training, as the computation that happens for each instance is conditioned on the input/output structures, and hence cannot be naïvely grouped together as a batch. This lack of parallelism is one of the major hurdles that prevent these representations from wider adoption practically (e.g., neural machine translation), as many natural language processing tasks currently require the ability to scale up to very large training corpora in order to reach state-of-the-art performance.

We seek to advance the state-of-the-art of this problem by proposing a parallelization scheme for one such network architecture, the Stack Long Short-Term Memory (StackLSTM) proposed in \citet{DBLP:conf/acl/DyerBLMS15}. This architecture has been successfully applied to dependency parsing \cite{DBLP:conf/acl/DyerBLMS15,DBLP:conf/naacl/DyerKBS16,DBLP:journals/coling/BallesterosDGS17} and syntax-aware neural machine translation \cite{DBLP:conf/acl/EriguchiTC17} in the previous research literature, but none of these research results were produced with minibatched training. We show that our parallelization scheme is feasible in practice by showing that it scales up near-linearly with increasing batch size, while reproducing a set of results reported in \cite{DBLP:journals/coling/BallesterosDGS17}.
\section{StackLSTM}

StackLSTM \cite{DBLP:conf/acl/DyerBLMS15} is an LSTM architecture \cite{DBLP:journals/neco/HochreiterS97} augmented with a stack $\mathcal{H}$ that stores some of the hidden states built in the past. Unlike traditional LSTMs that always build state $h_t$ from $h_{t-1}$, the states of StackLSTM are built from the head of the state stack $\mathcal{H}$, maintained by a stack top pointer $p(\mathcal{H})$. At each time step, StackLSTM takes a real-valued input vector together with an additional discrete operation on the stack, which determines what computation needs to be conducted and how the stack top pointer should be updated. Throughout this section, we index the input vector (e.g. word embeddings) $x_t$ using the time step $t$ it is fed into the network, and hidden states in the stack $h_j$ using their position $j$ in the stack $\mathcal{H}$, $j$ being defined as the 0-base index starting from the stack bottom.

The set of input discrete actions typically contains at least \texttt{Push} and \texttt{Pop} operations. When these operations are taken as input, the corresponding computations on the StackLSTM are listed below:\footnote{To simplify the presentation, we omitted the updates on cell states, because in practice the operations performed on cell states and hidden states are the same.}

\begin{table*}[t]
\centering
\begin{tabular}{lllll}
\toprule
\textbf{Transition Systems}   & \textbf{Transition Op} & \textbf{Stack Op} & \textbf{Buffer Op} & \textbf{Composition Op} \\ 
\midrule
\multirow{3}{*}{Arc-Standard} & Shift                  & push              & pop                & none                    \\
                              & Left-Arc               & pop, pop, push    & hold               & $S1 \leftarrow g(S0, S1)$             \\
                              & Right-Arc              & pop               & hold               & $S1 \leftarrow g(S1, S0)$             \\ \midrule
\multirow{4}{*}{Arc-Eager}    & Shift                  & push              & pop                & none                    \\
                              & Reduce                 & pop               & hold               & none                    \\
                              & Left-Arc               & pop               & hold               & $B0 \leftarrow g(B0, S0)$             \\
                              & Right-Arc              & push              & pop                & $B0 \leftarrow g(S0, B0)$             \\ \midrule
\multirow{3}{*}{Arc-Hybrid}   & Shift                  & push              & pop                & none                    \\
                              & Left-Arc               & pop               & hold               & $B0 \leftarrow g(B0, S0)$             \\
                              & Right-Arc              & pop               & hold               & $S1 \leftarrow g(S1, S0)$             \\ \bottomrule
\end{tabular}
\caption{Correspondence between transition operations and stack/buffer operations for StackLSTM, where $g$ denotes the composition function as proposed by \cite{DBLP:conf/acl/DyerBLMS15}. S0 and B0 refers to the token-level representation corresponding to the top element of the stack and buffer, while S1 and B1 refers to those that are second to the top. We use a different notation here to avoid confusion with the states in StackLSTM, which represent non-local information beyond token-level.}
\label{tab:stack-op}
\end{table*}

\vspace{-2mm}
\begin{itemize} \itemsep -2pt
\item \texttt{Push}: read previous hidden state $h_{p(\mathcal{H})}$, perform LSTM forward computation with $x_t$ and $h_{p(\mathcal{H})}$, write new hidden state to $h_{p(\mathcal{H}) + 1}$
, update stack top pointer with $p(\mathcal{H}) \leftarrow p(\mathcal{H}) + 1$.
\item \texttt{Pop}: update stack top pointer with $p(\mathcal{H}) \leftarrow p(\mathcal{H}) - 1$.
\end{itemize}
\vspace{-2mm}


Reflecting on the aforementioned discussion on parallelism, one should notice that StackLSTM falls into the category of neural network architectures that is difficult to perform minibatched training. This is caused by the fact that the computation performed by StackLSTM at each time step is dependent on the discrete input actions. The following section proposes a solution to this problem.

\section{Parallelizable StackLSTM\label{sec:p-stacklstm}}



Continuing the formulation in the previous section, we will start by discussing our proposed solution under the case where the set of discrete actions contains only \texttt{Push} and \texttt{Pop} operations; we then move on to discussion of the applicability of our proposed solution to the transition systems that are used for building representations for dependency trees.

The first modification we perform to the \texttt{Push} and \texttt{Pop} operations above is to unify the pointer update of these operations as $p(\mathcal{H}) \leftarrow p(\mathcal{H}) + op$, where $op$ is the input discrete operation that could either take the value +1 or -1 for \texttt{Push} and \texttt{Pop} operation.  After this modification, we came to the following observations:

\begin{observation}\label{obs:subset}
The computation performed for \texttt{Pop} operation is a subset of \texttt{Push} operation.
\end{observation}

Now, what remains to homogenize \texttt{Push} and \texttt{Pop} operations is conducting the extra computations needed for \texttt{Push} operation when \texttt{Pop} is fed in as well, while guaranteeing the correctness of the resulting hidden state both in the current time step and in the future. The next observation points out a way for this guarantee:

\begin{observation}\label{obs:locality}
In a StackLSTM, given the current stack top pointer position $p(\mathcal{H})$, any hidden state $h_i$ where $i > p(\mathcal{H})$ will not be read until it is overwritten by a \texttt{Push} operation.
\end{observation}

What follows from this observation is the guarantee that we can always safely overwrite hidden states $h_i$ that are indexed higher than the current stack top pointer, because it is known that any read operation on these states will happen after another overwrite. This allows us to {\it do the extra computation anyway} when \texttt{Pop} operation is fed, because the extra computation, especially updating $h_{p(\mathcal{H}) + 1}$, will not harm the validity of the hidden states at any time step.

Algorithm \ref{algo1} gives the final forward computation for the Parallelizable StackLSTM. Note that this algorithm does not contain any if-statements that depends on stack operations and hence is homogeneous when grouped into batches that are consisted of multiple operations trajectories.

\RestyleAlgo{ruled}
\begin{algorithm}
 \KwIn{input vector $x_t$\\\quad\quad discrete stack operation $op$}
 \KwOut{current top hidden state $h_{p(\mathcal{H})}$}
 $h\_prev$ $\leftarrow h_{p(\mathcal{H})}$\;
 $h \leftarrow \text{LSTM}(x_t, h\_prev)$\;
 $h_{p(\mathcal{H}) + 1}\leftarrow h$\;
 $p(\mathcal{H}) \leftarrow p(\mathcal{H}) + op$\;
 \Return{$h_{p(\mathcal{H})}$}\;
 \caption{Forward Computation for Parallelizable StackLSTM\label{algo1}\vspace{0.1cm}}
\end{algorithm}







In transition systems \cite{DBLP:journals/coling/Nivre08,DBLP:conf/acl/KuhlmannGS11} used in real tasks (e.g., transition-based parsing) as shown in Table \ref{tab:stack-op}, it should be noted that more than \texttt{push} and \texttt{pop} operations are needed for the StackLSTM. Fortunately, for Arc-Eager and Arc-Hybrid transition systems, we can simply add a \texttt{hold} operation, which is denoted by value 0 for the discrete operation input. For that reason, we will focus on parallelization of these two transition systems for this paper. It should be noted that both observations discussed above are still valid after adding the \texttt{hold} operation.

\section{Experiments}

\subsection{Setup}


We implemented\footnote{\tt https://github.com/shuoyangd/hoolock} the architecture described above in PyTorch  \cite{paszke2017automatic}. We implemented the batched stack as a float tensor wrapped in a non-leaf variable, thus enabling in-place operations on that variable. At each time step, the batched stack is queried/updated with a batch of stack head positions represented by an integer vector, an operation made possible by \texttt{gather} operation and advanced indexing. Due to this implementation choice, the stack size has to be determined at initialization time and cannot be dynamically grown. Nonetheless, a fixed stack size of 150 works for all the experiments we conducted. 

We use the dependency parsing task to evaluate the correctness and the scalability of our method. For comparison with previous work, we follow the architecture introduced in \citet{DBLP:conf/acl/DyerBLMS15,DBLP:journals/coling/BallesterosDGS17} and chose the Arc-Hybrid transition system for comparison with previous work. We follow the data setup in \citet{DBLP:conf/emnlp/ChenM14,DBLP:conf/acl/DyerBLMS15,DBLP:journals/coling/BallesterosDGS17} and use Stanford Dependency Treebank \cite{DBLP:conf/lrec/MarneffeMM06} for dependency parsing, and we extract the Arc-Hybrid static oracle using the code associated with \citet{DBLP:conf/acl/QiM17}. The part-of-speech (POS) tags are generated with Stanford POS-tagger \cite{DBLP:conf/naacl/ToutanovaKMS03} with a test set accuracy of 97.47\%. We use exactly the same pre-trained English word embedding as \citet{DBLP:conf/acl/DyerBLMS15}.

We use Adam \cite{DBLP:journals/corr/KingmaB14} as the optimization algorithm. Following \citet{DBLP:journals/corr/GoyalDGNWKTJH17}, we apply linear warmup to the learning rate with an initial value of $\tau = 5\times 10^{-4}$ and total epoch number of 5. The target learning rate is set by $\tau$ multiplied by batch size, but capped at 0.02 because we find Adam to be unstable beyond that learning rate. After warmup, we reduce the learning rate by half every time there is no improvement for loss value on the development set (\texttt{ReduceLROnPlateau}). We clip all the gradient norms to 5.0 and apply a L$_2$-regularization with weight $1\times 10^{-6}$.

We started with the hyper-parameter choices in \citet{DBLP:conf/acl/DyerBLMS15} but made some modifications based on the performance on the development set: we use hidden dimension 200 for all the LSTM units, 200 for the parser state representation before the final softmax layer, and embedding dimension 48 for the action embedding.

We use Tesla K80 for all the experiments, in order to compare with \citet{DBLP:conf/nips/NeubigGD17,DBLP:conf/acl/DyerBLMS15}. We also use the same hyper-parameter setting as \citet{DBLP:conf/acl/DyerBLMS15} for speed comparison experiments. All the speeds are measured by running through one training epoch and averaging.

\subsection{Results}

\begin{figure}[t]
\includegraphics[width=0.45\textwidth]{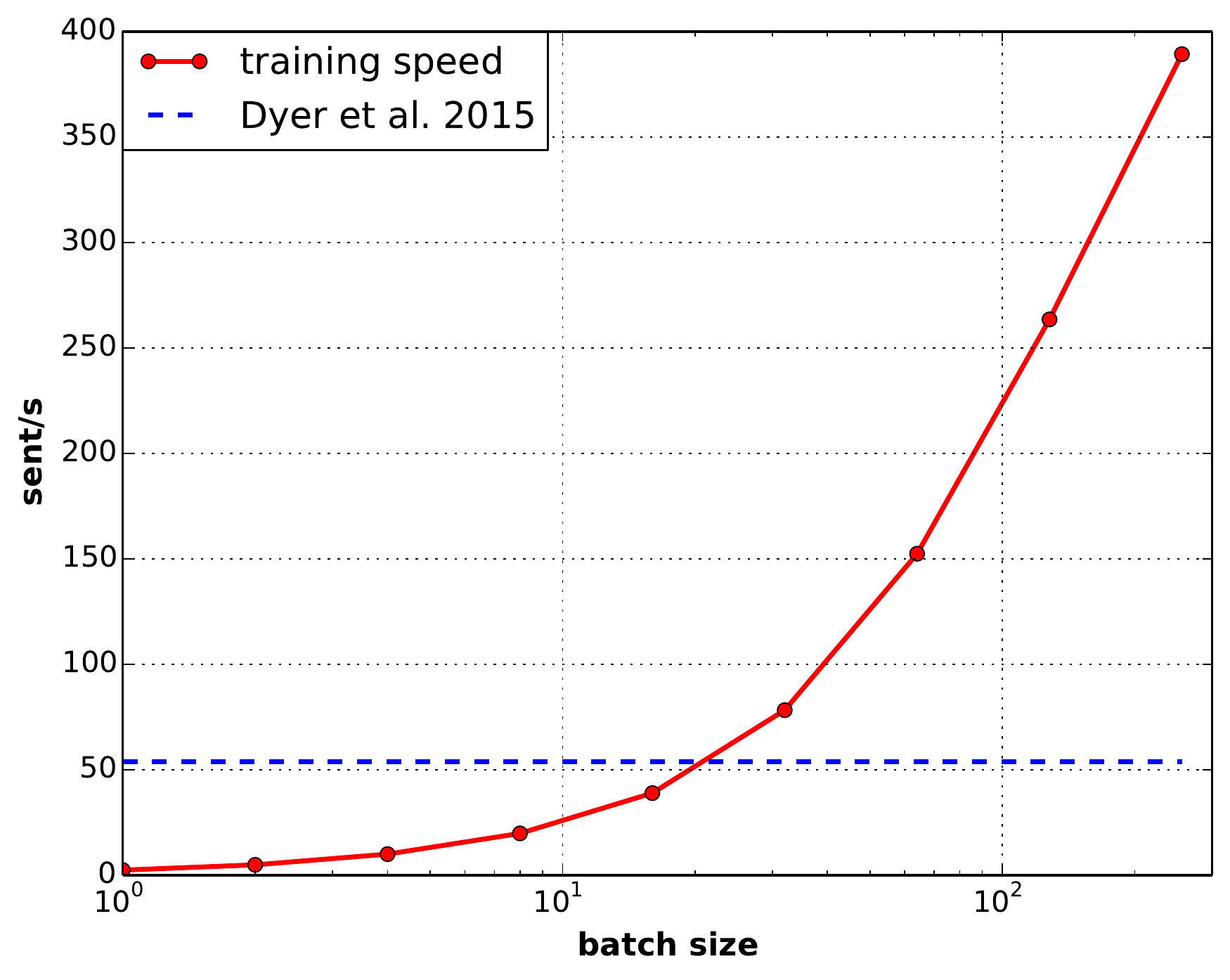}
\caption{Training speed at different batch size. Note that the $x$-axis is in log-scale in order to show all the data points properly.\label{fig-speedup}}
\end{figure}

\begin{table}[htbp]
\centering
\begin{tabular}{lllll}
\toprule
\multirow{2}{*}{$b$} & \multicolumn{2}{l}{\textbf{dev}} & \multicolumn{2}{l}{\textbf{test}} \\ \cline{2-5} 
                     & UAS        & LAS        & UAS         & LAS        \\ \midrule
1*                   & 92.50*     & 89.79*     & 92.10*      & 89.61*     \\ \midrule
8                    & 92.93      & 90.42      & 92.54       & 90.11      \\ \midrule
16                   & 92.62      & 90.19      & 92.53       & 90.13      \\ \midrule
32                   & 92.43      & 89.89      & 92.31       & 89.94      \\ \midrule
64                   & 92.53      & 90.04      & 92.22       & 89.73      \\ \midrule
128                  & 92.39      & 89.73      & 92.55       & 90.02      \\ \midrule
256                  & 92.15      & 89.46      & 91.99       & 89.43      \\ \bottomrule
\end{tabular}
\caption{Dependency parsing result with various training batch size $b$ and without composition function. The results marked with asterisks were reported in the \citet{DBLP:journals/coling/BallesterosDGS17}.\label{tab-parsing}}
\end{table}

Figure \ref{fig-speedup} shows the training speed at different batch sizes up to 256.\footnote{At batch size of 512, the longest sentence in the training data cannot be fit onto the GPU.} The speed-up of our model is close to linear, which means there is very little overhead associated with our batching scheme. Quantitatively, according to Amdahl's Law \cite{DBLP:conf/afips/Amdahl67}, the proportion of parallelized computations is 99.92\% at batch size 64. We also compared our implementation with the implementation that comes with \citet{DBLP:conf/acl/DyerBLMS15}, which is implemented in C++ with DyNet \cite{DBLP:journals/corr/NeubigDGMAABCCC17}. DyNet is known to be very optimized for CPU computations and hence their implementation is reasonably fast even without batching and GPU acceleration, as shown in Figure \ref{fig-speedup}.\footnote{Measured on one core of an Intel Xeon E7-4830 CPU.} But we would like to point out that we focus on the speed-up we are able to obtain rather than the absolute speed, and that our batching scheme is framework-universal and superior speed might be obtained by combining our scheme with alternative frameworks or languages (for example, the torch C++ interface).

The dependency parsing results are shown in Table \ref{tab-parsing}. Our implementation is able to yield better test set performance than that reported in \citet{DBLP:journals/coling/BallesterosDGS17} for all batch size configurations except 256, where we observe a modest performance loss. Like \citet{DBLP:journals/corr/GoyalDGNWKTJH17,DBLP:journals/corr/KeskarMNST16,DBLP:journals/corr/abs-1804-07612}, we initially observed more significant test-time performance deterioration (around $1\%$ absolute difference) for models trained without learning rate warmup, and concurring with the findings in \citet{DBLP:journals/corr/GoyalDGNWKTJH17}, we find warmup very helpful for stabilizing large-batch training. We did not run experiments with batch size below 8 as they are too slow due to Python's inherent performance issue.


\section{Related Work}

DyNet has support for automatic minibatching \cite{DBLP:conf/nips/NeubigGD17}, which figures out what computation is able to be batched by traversing the computation graph to find homogeneous computations.
While we cannot directly compare with that framework's automatic batching solution for StackLSTM\footnote{This is due to the fact that DyNet automatic batching cannot handle graph structures that depends on runtime input values, which is the case in StackLSTM.}, we can draw a loose comparison to the results reported in that paper for BiLSTM transition-based parsing \cite{DBLP:journals/corr/KiperwasserG16a}. Comparing batch size of 64 to batch size of 1, they obtained a 3.64x speed-up on CPU and 2.73x speed-up on Tesla K80 GPU, while our architecture-specific manual batching scheme obtained 60.8x speed-up. The main reason for this difference is that their graph-traversing automatic batching scheme carries a much larger overhead compared to our manual batching approach.

Another toolkit that supports automatic minibatching is Matchbox\footnote{\tt https://github.com/salesforce/matchbox}, which operates by analyzing the single-instance model definition and deterministically convert the operations into their minibatched counterparts. While such mechanism eliminated the need to traverse the whole computation graph, it cannot homogenize the operations in each branch of \texttt{if}. Instead, it needs to perform each operation separately and apply masking on the result, while our method does not require any masking. Unfortunately we are also not able to compare with the toolkit at the time of this work as it lacks support for several operations we need.

Similar to the spirit of our work, \citet{DBLP:conf/acl/BowmanGRGMP16} attempted to parallelize StackLSTM by using \textit{Thin-stack}, a data structure that reduces the space complexity by storing all the intermediate stack top elements in a tensor and use a queue to control element access. However, thanks to PyTorch, our implementation is not directly dependent on the notion of Thin-stack. Instead, when an element is popped from the stack, we simply shift the stack top pointer and potentially re-write the corresponding sub-tensor later. In other words, there is no need for us to directly maintain all the intermediate stack top elements, because in PyTorch, when the element in the stack is re-written, its underlying sub-tensor will not be destructed as there are still nodes in the computation graph that point to it. Hence, when performing back-propagation, the gradient is still able to flow back to the elements that are previously popped from the stack and their respective precedents. Hence, we are also effectively storing all the intermediate stack top elements only once. Besides, \citet{DBLP:conf/acl/BowmanGRGMP16} didn't attempt to eliminate the conditional branches in the StackLSTM algorithm, which is the main algorithmic contribution of this work.
\section{Conclusion}

We propose a parallelizable version of StackLSTM that is able to fully exploit the GPU parallelism by performing minibatched training. Empirical results show that our parallelization scheme yields comparable performance to previous work, and our method scales up very linearly with the increasing batch size.

Because our parallelization scheme is based on the observation made in section \ref{obs:subset}, we cannot incorporate batching for neither Arc-Standard transition system nor the token-level composition function proposed in \citet{DBLP:conf/acl/DyerBLMS15} efficiently yet. 
We leave the parallelization of these architectures to future work.

Our parallelization scheme makes it feasible to run large-data experiments for various tasks that requires large training data to perform well, such as RNNG-based syntax-aware neural machine translation \cite{DBLP:conf/acl/EriguchiTC17}.

\section*{Acknowledgement}
The authors would like to thank Peng Qi for helping with data preprocessing and James Bradbury for helpful technical discussions. This material is based upon work supported in part by the DARPA LORELEI and IARPA MATERIAL programs.

\bibliography{final}

\begin{thebibliography}{22}
\expandafter\ifx\csname natexlab\endcsname\relax\def\natexlab#1{#1}\fi

\bibitem[{Amdahl(1967)}]{DBLP:conf/afips/Amdahl67}
Gene~M. Amdahl. 1967.
\newblock Validity of the single processor approach to achieving large scale
  computing capabilities.
\newblock In \emph{American Federation of Information Processing Societies:
  Proceedings of the {AFIPS} '67 Spring Joint Computer Conference, April 18-20,
  1967, Atlantic City, New Jersey, {USA}}, pages 483--485.

\bibitem[{Ballesteros et~al.(2017)Ballesteros, Dyer, Goldberg, and
  Smith}]{DBLP:journals/coling/BallesterosDGS17}
Miguel Ballesteros, Chris Dyer, Yoav Goldberg, and Noah~A. Smith. 2017.
\newblock Greedy transition-based dependency parsing with stack lstms.
\newblock \emph{Computational Linguistics}, 43(2):311--347.

\bibitem[{Bowman et~al.(2016)Bowman, Gauthier, Rastogi, Gupta, Manning, and
  Potts}]{DBLP:conf/acl/BowmanGRGMP16}
Samuel~R. Bowman, Jon Gauthier, Abhinav Rastogi, Raghav Gupta, Christopher~D.
  Manning, and Christopher Potts. 2016.
\newblock A fast unified model for parsing and sentence understanding.
\newblock In \emph{Proceedings of the 54th Annual Meeting of the Association
  for Computational Linguistics, {ACL} 2016, August 7-12, 2016, Berlin,
  Germany, Volume 1: Long Papers}.

\bibitem[{Chen and Manning(2014)}]{DBLP:conf/emnlp/ChenM14}
Danqi Chen and Christopher~D. Manning. 2014.
\newblock A fast and accurate dependency parser using neural networks.
\newblock In \emph{Proceedings of the 2014 Conference on Empirical Methods in
  Natural Language Processing, {EMNLP} 2014, October 25-29, 2014, Doha, Qatar,
  {A} meeting of SIGDAT, a Special Interest Group of the {ACL}}, pages
  740--750.

\bibitem[{Dyer et~al.(2015)Dyer, Ballesteros, Ling, Matthews, and
  Smith}]{DBLP:conf/acl/DyerBLMS15}
Chris Dyer, Miguel Ballesteros, Wang Ling, Austin Matthews, and Noah~A. Smith.
  2015.
\newblock Transition-based dependency parsing with stack long short-term
  memory.
\newblock In \emph{Proceedings of the 53rd Annual Meeting of the Association
  for Computational Linguistics and the 7th International Joint Conference on
  Natural Language Processing of the Asian Federation of Natural Language
  Processing, {ACL} 2015, July 26-31, 2015, Beijing, China, Volume 1: Long
  Papers}, pages 334--343.

\bibitem[{Dyer et~al.(2016)Dyer, Kuncoro, Ballesteros, and
  Smith}]{DBLP:conf/naacl/DyerKBS16}
Chris Dyer, Adhiguna Kuncoro, Miguel Ballesteros, and Noah~A. Smith. 2016.
\newblock Recurrent neural network grammars.
\newblock In \emph{{NAACL} {HLT} 2016, The 2016 Conference of the North
  American Chapter of the Association for Computational Linguistics: Human
  Language Technologies, San Diego California, USA, June 12-17, 2016}, pages
  199--209.

\bibitem[{Eriguchi et~al.(2017)Eriguchi, Tsuruoka, and
  Cho}]{DBLP:conf/acl/EriguchiTC17}
Akiko Eriguchi, Yoshimasa Tsuruoka, and Kyunghyun Cho. 2017.
\newblock Learning to parse and translate improves neural machine translation.
\newblock In \emph{Proceedings of the 55th Annual Meeting of the Association
  for Computational Linguistics, {ACL} 2017, Vancouver, Canada, July 30 -
  August 4, Volume 2: Short Papers}, pages 72--78.

\bibitem[{Goyal et~al.(2017)Goyal, Doll{\'{a}}r, Girshick, Noordhuis,
  Wesolowski, Kyrola, Tulloch, Jia, and
  He}]{DBLP:journals/corr/GoyalDGNWKTJH17}
Priya Goyal, Piotr Doll{\'{a}}r, Ross~B. Girshick, Pieter Noordhuis, Lukasz
  Wesolowski, Aapo Kyrola, Andrew Tulloch, Yangqing Jia, and Kaiming He. 2017.
\newblock Accurate, large minibatch {SGD:} training imagenet in 1 hour.
\newblock \emph{CoRR}, abs/1706.02677.

\bibitem[{Hochreiter and Schmidhuber(1997)}]{DBLP:journals/neco/HochreiterS97}
Sepp Hochreiter and J{\"{u}}rgen Schmidhuber. 1997.
\newblock Long short-term memory.
\newblock \emph{Neural Computation}, 9(8):1735--1780.

\bibitem[{Keskar et~al.(2016)Keskar, Mudigere, Nocedal, Smelyanskiy, and
  Tang}]{DBLP:journals/corr/KeskarMNST16}
Nitish~Shirish Keskar, Dheevatsa Mudigere, Jorge Nocedal, Mikhail Smelyanskiy,
  and Ping Tak~Peter Tang. 2016.
\newblock On large-batch training for deep learning: Generalization gap and
  sharp minima.
\newblock \emph{CoRR}, abs/1609.04836.

\bibitem[{Kingma and Ba(2014)}]{DBLP:journals/corr/KingmaB14}
Diederik~P. Kingma and Jimmy Ba. 2014.
\newblock Adam: {A} method for stochastic optimization.
\newblock \emph{CoRR}, abs/1412.6980.

\bibitem[{Kiperwasser and Goldberg(2016)}]{DBLP:journals/corr/KiperwasserG16a}
Eliyahu Kiperwasser and Yoav Goldberg. 2016.
\newblock Simple and accurate dependency parsing using bidirectional {LSTM}
  feature representations.
\newblock \emph{CoRR}, abs/1603.04351.

\bibitem[{Kuhlmann et~al.(2011)Kuhlmann, G{\'{o}}mez{-}Rodr{\'{\i}}guez, and
  Satta}]{DBLP:conf/acl/KuhlmannGS11}
Marco Kuhlmann, Carlos G{\'{o}}mez{-}Rodr{\'{\i}}guez, and Giorgio Satta. 2011.
\newblock Dynamic programming algorithms for transition-based dependency
  parsers.
\newblock In \emph{The 49th Annual Meeting of the Association for Computational
  Linguistics: Human Language Technologies, Proceedings of the Conference,
  19-24 June, 2011, Portland, Oregon, {USA}}, pages 673--682.

\bibitem[{de~Marneffe et~al.(2006)de~Marneffe, MacCartney, and
  Manning}]{DBLP:conf/lrec/MarneffeMM06}
Marie{-}Catherine de~Marneffe, Bill MacCartney, and Christopher~D. Manning.
  2006.
\newblock Generating typed dependency parses from phrase structure parses.
\newblock In \emph{Proceedings of the Fifth International Conference on
  Language Resources and Evaluation, {LREC} 2006, Genoa, Italy, May 22-28,
  2006.}, pages 449--454.

\bibitem[{Masters and Luschi(2018)}]{DBLP:journals/corr/abs-1804-07612}
Dominic Masters and Carlo Luschi. 2018.
\newblock Revisiting small batch training for deep neural networks.
\newblock \emph{CoRR}, abs/1804.07612.

\bibitem[{Neubig et~al.(2017{\natexlab{a}})Neubig, Dyer, Goldberg, Matthews,
  Ammar, Anastasopoulos, Ballesteros, Chiang, Clothiaux, Cohn, Duh, Faruqui,
  Gan, Garrette, Ji, Kong, Kuncoro, Kumar, Malaviya, Michel, Oda, Richardson,
  Saphra, Swayamdipta, and Yin}]{DBLP:journals/corr/NeubigDGMAABCCC17}
Graham Neubig, Chris Dyer, Yoav Goldberg, Austin Matthews, Waleed Ammar,
  Antonios Anastasopoulos, Miguel Ballesteros, David Chiang, Daniel Clothiaux,
  Trevor Cohn, Kevin Duh, Manaal Faruqui, Cynthia Gan, Dan Garrette, Yangfeng
  Ji, Lingpeng Kong, Adhiguna Kuncoro, Gaurav Kumar, Chaitanya Malaviya, Paul
  Michel, Yusuke Oda, Matthew Richardson, Naomi Saphra, Swabha Swayamdipta, and
  Pengcheng Yin. 2017{\natexlab{a}}.
\newblock Dynet: The dynamic neural network toolkit.
\newblock \emph{CoRR}, abs/1701.03980.

\bibitem[{Neubig et~al.(2017{\natexlab{b}})Neubig, Goldberg, and
  Dyer}]{DBLP:conf/nips/NeubigGD17}
Graham Neubig, Yoav Goldberg, and Chris Dyer. 2017{\natexlab{b}}.
\newblock On-the-fly operation batching in dynamic computation graphs.
\newblock In \emph{Advances in Neural Information Processing Systems 30: Annual
  Conference on Neural Information Processing Systems 2017, 4-9 December 2017,
  Long Beach, CA, {USA}}, pages 3974--3984.

\bibitem[{Nivre(2008)}]{DBLP:journals/coling/Nivre08}
Joakim Nivre. 2008.
\newblock Algorithms for deterministic incremental dependency parsing.
\newblock \emph{Computational Linguistics}, 34(4):513--553.

\bibitem[{Paszke et~al.(2017)Paszke, Gross, Chintala, Chanan, Yang, DeVito,
  Lin, Desmaison, Antiga, and Lerer}]{paszke2017automatic}
Adam Paszke, Sam Gross, Soumith Chintala, Gregory Chanan, Edward Yang, Zachary
  DeVito, Zeming Lin, Alban Desmaison, Luca Antiga, and Adam Lerer. 2017.
\newblock Automatic differentiation in pytorch.
\newblock In \emph{NIPS-W}.

\bibitem[{Qi and Manning(2017)}]{DBLP:conf/acl/QiM17}
Peng Qi and Christopher~D. Manning. 2017.
\newblock Arc-swift: {A} novel transition system for dependency parsing.
\newblock In \emph{Proceedings of the 55th Annual Meeting of the Association
  for Computational Linguistics, {ACL} 2017, Vancouver, Canada, July 30 -
  August 4, Volume 2: Short Papers}, pages 110--117.

\bibitem[{Socher et~al.(2011)Socher, Lin, Ng, and
  Manning}]{DBLP:conf/icml/SocherLNM11}
Richard Socher, Cliff~Chiung{-}Yu Lin, Andrew~Y. Ng, and Christopher~D.
  Manning. 2011.
\newblock Parsing natural scenes and natural language with recursive neural
  networks.
\newblock In \emph{Proceedings of the 28th International Conference on Machine
  Learning, {ICML} 2011, Bellevue, Washington, USA, June 28 - July 2, 2011},
  pages 129--136.

\bibitem[{Toutanova et~al.(2003)Toutanova, Klein, Manning, and
  Singer}]{DBLP:conf/naacl/ToutanovaKMS03}
Kristina Toutanova, Dan Klein, Christopher~D. Manning, and Yoram Singer. 2003.
\newblock Feature-rich part-of-speech tagging with a cyclic dependency network.
\newblock In \emph{Human Language Technology Conference of the North American
  Chapter of the Association for Computational Linguistics, {HLT-NAACL} 2003,
  Edmonton, Canada, May 27 - June 1, 2003}.

\end{thebibliography}
\bibliographystyle{acl_natbib_nourl}

\appendix

\end{document}